\documentclass[conference]{IEEEtran}
\IEEEoverridecommandlockouts
\usepackage{amsmath,amsfonts}
\usepackage{algorithmic}
\usepackage{algorithm}
\usepackage{array}
\usepackage{textcomp}
\usepackage{stfloats}
\usepackage{url}
\usepackage{verbatim}
\usepackage{subfigure}
\usepackage{graphicx,subcaption}
\usepackage{cite}
\usepackage[numbers,sort&compress]{natbib}

\usepackage{xcolor}
\usepackage{multirow}
\usepackage{makecell}
\usepackage{enumitem}
\usepackage{bm}
\usepackage{amssymb}
\usepackage{amsthm}

\usepackage{soul}

\theoremstyle{plain}

\allowdisplaybreaks[4]

\begin{document}

\title{Point Cloud Compression with Implicit Neural Representations: A Unified Framework}

\author{Hongning Ruan, Yulin Shao, Qianqian Yang, Liang Zhao, Dusit Niyato
\thanks{H. Ruan, Q. Yang and L. Zhao are with the Department of Information Science and Electronic Engineering, Zhejiang University (e-mails: \{rhohenning,qianqianyang20, lzhao2020\}@zju.edu.cn).

Y. Shao is with the State Key Laboratory of Internet of Things for Smart City and the Department of Electrical and Computer Engineering, University of Macau, Macau S.A.R. (e-mail: ylshao@um.edu.mo).

D. Niyato is with the School of Computer Science and Engineering, Nanyang Technological University, Singapore(e-mail: dniyato@ntu.edu.cn)

}}

\maketitle

\begin{abstract}
Point clouds have become increasingly vital across various applications thanks to their ability to realistically depict 3D objects and scenes. Nevertheless, effectively compressing unstructured, high-precision point cloud data remains a significant challenge. In this paper, we present a pioneering point cloud compression framework capable of handling both geometry and attribute components. 
Unlike traditional approaches and existing learning-based methods, our framework utilizes two coordinate-based neural networks to implicitly represent a voxelized point cloud. The first network generates the occupancy status of a voxel, while the second network determines the attributes of an occupied voxel. To tackle an immense number of voxels within the volumetric space, we partition the space into smaller cubes and focus solely on voxels within non-empty cubes.
By feeding the coordinates of these voxels into the respective networks, we reconstruct the geometry and attribute components of the original point cloud. The neural network parameters are further quantized and compressed. 
Experimental results underscore the superior performance of our proposed method compared to the octree-based approach employed in the latest G-PCC standards. Moreover, our method exhibits high universality when contrasted with existing learning-based techniques.
\end{abstract}

\begin{IEEEkeywords}
Point cloud compression, neural implicit representation, neural network compression.
\end{IEEEkeywords}

\section{Introduction}\label{sec:I}
Point clouds have become a ubiquitous format for representing 3D objects and scenes, finding applications in diverse fields like autonomous driving, augmented reality/virtual reality (AR/VR), digital twin, and robotics \cite{graziosi2020pccstd,wang2021learnedpcgc,PCmag}. Essentially, a point cloud comprises a multitude of 3D points scattered throughout volumetric space, each defined by its geometric location. These points' coordinates can be quantized into integer values, giving rise to voxel grids in the space, a process known as voxelization. Alongside geometry, each point in a point cloud is accompanied by corresponding attributes such as color, normal, and reflectance.
Advancements in sensing technologies have enabled the capture of large-scale point clouds with high-resolution spatial location and attribute information. However, the sheer size and unstructured nature of point cloud data require significant memory for storage or bandwidth for transmission, underscoring the need for more efficient compression approaches \cite{graziosi2020pccstd}.

The standardization of point cloud compression (PCC) began under the Moving Picture Expert Group (MPEG) in 2017 and was finalized in 2020, resulting in two distinct approaches: video-based PCC (V-PCC) and geometry-based PCC (G-PCC) \cite{graziosi2020pccstd}. Both methods rely on traditional representations of point cloud data, such as octrees, triangle meshes, and 3D-to-2D projection.
In contrast, recent research efforts \cite{wang2021learnedpcgc,PCmag,wang2021pcgcv2,liu2022pcgformer, wang2022sparsepcgc,wang2022sparsepcac,SEPT} have explored the application of deep learning techniques in point cloud compression. Most of the existing works focus on the compression of either geometry or attribute. They utilize the autoencoder architecture to process the point cloud directly, where an encoder transforms the input point cloud into latent representation and a decoder reconstructs the input. Despite demonstrating performance gains over traditional methods, these learning-based approaches require extensive point cloud datasets for network training. Moreover, as highlighted in \cite{alexiou2020unified}, the selection of training data significantly influences network performance on testing data, necessitating improved generalization capabilities.

In recent years, deep neural networks (DNNs) have been employed to represent 3D objects and scenes implicitly, exemplified by NeRF \cite{mildenhall2021nerf}. Such networks learn continuous functions that take coordinates as input and output the corresponding features. This kind of method, known as neural implicit representations (NIR), offers a novel avenue for point cloud compression, with several relevant works already in existence.
For instance, NVFPCC \cite{hu2022nvfpcc} trains a convolutional neural network alongside input latent codes, with each code used to reconstruct a group of points. LVAC \cite{isik2022lvac} employs a coordinate-based network to represent point attributes, using latent vectors as local parameters. NIC \cite{pistilli2022nic} utilizes a single coordinate-based network for point cloud attribute representation, but with meta-learning to exploit prior knowledge and enhance compression efficiency. However, these studies predominantly focus on either geometry or attributes, leaving a gap for a unified framework that encompasses both aspects of compression, solely based on NIR.

In this paper, we introduce a novel framework for compressing both geometry and attribute components of point clouds. Our approach employs two coordinate-based DNNs to implicitly represent a voxelized point cloud. The first network categorizes each voxel as occupied or unoccupied based on input spatial coordinates, generating occupancy probabilities. A threshold is then applied to delineate between the two classes. By inputting voxel coordinates into the network and selecting occupied voxels, we reconstruct the geometry component of the original point cloud. The second network generates attributes for occupied voxels, taking spatial coordinates as input and outputting corresponding attributes (e.g., RGB colors). To encode the original point cloud, we fine-tune the two networks on the data and subsequently quantize and encode their parameters. During decoding, we first retrieve the decoded parameters and then use the two networks to reconstruct the point cloud.

Our main contributions are summarized as follows:

\begin{itemize}

\item  We introduce a unified framework for point cloud compression capable of processing both geometry and attribute components. Our method exhibits superior rate-distortion performance compared to the octree-based approach adopted by the latest G-PCC standard, whether for geometry compression alone or joint compression of geometry and attributes.

\item We pioneer the utilization of NIR in point cloud compression, opening up a new avenue for research in the field. We explore various strategies to enhance the representation ability and compressibility of neural networks.

\item Our proposed method demonstrates exceptional universality compared to other learning-based approaches. By training the networks on a single point cloud and without relying on specific datasets, our method can be applied to various point cloud data types.

\end{itemize}

\section{Proposed method}\label{sec:II}
\begin{figure*}[t]
\centering
\includegraphics[width=1.6\columnwidth]{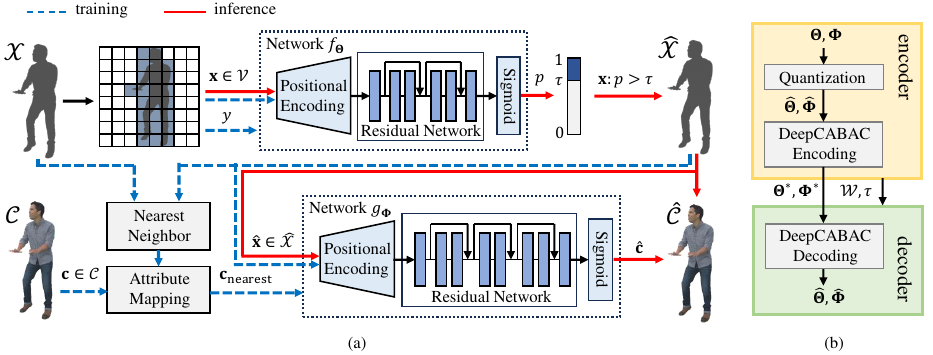}
\caption{Overview of the proposed method. (a) The training and inference procedure of the networks, indicated by dashed lines and solid lines respectively. (b) The coding procedure.}
\label{fig:overview}
\end{figure*}

\subsection{Overview}

We denote a point cloud as $\mathcal{P}=\{\mathcal{X}, \mathcal{C}\}$, where $\mathcal{X}$ represents the geometric components, i.e., the coordinates of all points, and $\mathcal{C}$ represents the attribute components, i.e., the corresponding color of each point. As shown in Fig. \ref{fig:overview}, to compress the point cloud, we first learn an implicit neural field, followed by quantization and encoding of neural parameters. These quantized parameters allow us to reconstruct a lossy version of the original point cloud, denoted by $\widehat{\mathcal{P}}=\{\widehat{\mathcal{X}}, \widehat{\mathcal{C}}\}$.

We employ a neural network $f_{\mathbf{\Theta}}$, where $\mathbf{\Theta}$ represents its parameters, to represent the geometry components of the original point cloud. The input to this network, denoted by $\mathbf{x}$, can be any voxel within the entire space, yielding an occupancy probability $p$ as output. Given that a significant portion of the space is typically vacant, we initially divide an entire space into smaller cubes and only consider voxels within non-empty cubes as inputs to the network. We denote the set of all non-empty cubes by $\mathcal{W}$, and the set of all voxels within these cubes by $\mathcal{V}$. During training, we optimize the network parameters to ensure that the output probability $p$ closely matches the ground-truth occupancy $y$. During inference, after obtaining $p$, we interpret the input voxel as occupied if $p$ exceeds a given threshold $\tau$. We then aggregate all occupied voxels to form the reconstructed geometry $\widehat{\mathcal{X}}$, given by $\widehat{\mathcal{X}}=\{\mathbf{x}:f_{\mathbf{\Theta}}(\mathbf{x})>\tau,\mathbf{x}\in\mathcal{V}\}$.

We utilize another neural network $g_{\mathbf{\Phi}}$, where $\mathbf{\Phi}$represents its parameters, to represent the attribute components of the original point cloud. When encoding attributes, we leverage the reconstructed geometry $\widehat{\mathcal{X}}$ as prior knowledge. The network takes an occupied voxel $\widehat{\mathbf{x}}\in\widehat{\mathcal{X}}$as input and generates its RGB color $\widehat{\mathbf{c}}$ as output. Throughout the training process, we adjust the network parameters to ensure that the predicted color for each input voxel closely aligns with the expected color. Since the reconstructed geometry $\widehat{\mathcal{X}}$ might not perfectly match the ground-truth geometry $\mathcal{X}$, we define the expected color of each voxel to match that of its nearest neighbor in the original geometry, aiming to minimize overall attribute distortion. During the inference phase, we input the occupied voxels into the network to obtain the reconstructed attributes, given by $\widehat{\mathcal{C}}=\{\widehat{\mathbf{c}}:\widehat{\mathbf{c}}=g_{\mathbf{\Phi}}(\widehat{\mathbf{x}}),\widehat{\mathbf{x}}\in\widehat{\mathcal{X}}\}$.

The encoder learns the network parameters $\mathbf{\Theta}$ and $\mathbf{\Phi}$ for a given point cloud trough the aforementioned training process. These parameters are then quantized followed by binarization, and are further encoded using DeepCABAC \cite{wiedemann2020deepcabac}. Additionally, $\mathcal{W}$ and $\tau$ are encoded as auxiliary information to be transmitted to the receiver, requiring only a small number of bits for representation. Subsequently, the decoder reconstructs the point cloud  $\widehat{\mathcal{P}}$ from the quantized parameters, alongside $\mathcal{V}$ and $\tau$, where $\mathcal{V}$ comprises all the voxels within $\mathcal{W}$, by following the aforementioned inference procedure.

\subsection{Volumetric Space Partitioning}

Assuming the original point cloud is voxelized with an $N$-bit resolution,  the volumetric space comprises $2^N\times 2^N\times 2^N$ voxels. Each voxel's location can be represented by a coordinate $\mathbf{x}\in \{0, \cdots, 2^N-1\}^3$. Due to a significant fraction of these voxels being empty, processing all voxels during either training or inference would be excessively time-consuming. To address this, we partition the space into $2^T\times 2^T\times 2^T$ cubes, each containing $2^{N-T}\times 2^{N-T}\times 2^{N-T}$ voxels. Similarly, the location of each cube can be represented by coordinate $\mathbf{w} \in \{0, \cdots, 2^T-1\}^3$. For a voxel $\mathbf{x}$, the cube containing it is given by $\mathbf{w}_{\mathbf{x}}=\mathrm{floor}(\mathbf{x}/2^{N-T})$. As previously mentioned, the set containing the coordinates of all non-empty cubes is denoted by $\mathcal{W}$. During both training and inference, we only consider voxels within these cubes, which can be given by $\mathcal{V}=\{\mathbf{x}:\mathbf{w}_{\mathbf{x}}\in\mathcal{W}\}$.

\subsection{Neural Network Structure}

\begin{figure}
\centering
\includegraphics[height=4.5cm]{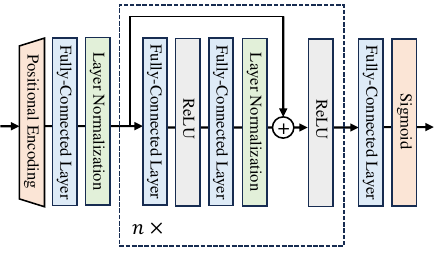}
\caption{The detailed structure for both networks.}
\label{fig:network_structure}
\end{figure}

We implement both functions $f_{\mathbf{\Theta}}, g_{\mathbf{\Phi}}$ using coordinate-based fully-connected networks. The detailed structure of both networks is depicted in Fig. \ref{fig:network_structure}. Each network consists of multiple residual blocks \cite{he2016residual} and two additional fully-connected layers, one at the input and the other at the output. Each residual block consists of two fully-connected layers, both employing ReLU activation. Additionally, layer normalization which normalizes each layer is performed on certain layers within the network. The final layer at the output utilizes the sigmoid function to ensure output values lie within the range $[0,1]$.

We employ positional encoding to transform the input coordinate into a higher-dimensional space, utilizing the encoding function proposed by NeRF \cite{mildenhall2021nerf}. This encoding function is defined as follows:
\begin{equation}\label{eq:positional_encoding}
\begin{aligned}
\gamma(\widetilde{\mathbf{x}})=(\widetilde{\mathbf{x}}, \sin(2^0\pi\widetilde{\mathbf{x}}), \cos(2^0\pi\widetilde{\mathbf{x}}), \cdots,\\
\sin(2^{L-1}\pi\widetilde{\mathbf{x}}), \cos(2^{L-1}\pi\widetilde{\mathbf{x}})),
\end{aligned}
\end{equation}
where $\widetilde{\mathbf{x}}$ is derived by normalizing the input coordinate $\mathbf{x}$ to the range of $[-1,1]$, and $L$ is the number of different frequencies. This positional encoding maps each coordinate component from a single scalar to a vector of length $(2L+1)$, thereby incorporating more high-frequency variations than the original input.

\subsection{Quantization and Coding}

The network parameters are quantized using a given step size $\Delta$, i.e., $k=\mathrm{round}(q/\Delta)$ for every parameter $q$. Subsequently, these quantized parameters are encoded using the binarization and encoding method proposed by DeepCABAC \cite{wiedemann2020deepcabac}. This method efficiently encodes zero parameters using very few bits, enabling high compression ratios particularly on sparsified networks.

\subsection{Loss Function}

We optimize the two networks $f_{\mathbf{\Theta}}$ and $g_{\mathbf{\Phi}}$ by minimizing the following loss functions respectively, i.e.,
\begin{gather}
\mathcal{L}_f=\dfrac{1}{|\mathcal{B}_f|}\sum_{\mathbf{x}\in\mathcal{B}_f}D_f(\mathbf{x})+\dfrac{\lambda_f}{|\mathcal{X}|}\|\mathbf{\Theta}\|_1,\\
\mathcal{L}_g=\dfrac{1}{|\mathcal{B}_g|}\sum_{\widehat{\mathbf{x}}\in\mathcal{B}_g}D_g(\widehat{\mathbf{x}})+\dfrac{\lambda_g}{|\mathcal{X}|}\|\mathbf{\Phi}\|_1,
\end{gather}
where $D_f(\cdot)$ and $D_g(\cdot)$ denote the geometry and attribute distortion loss functions. $\mathcal{B}_f$ and $\mathcal{B}_g$ denote training batches. 
\subsubsection{Geometry Distortion}
The geometry distortion loss function is defined as $\alpha$-balanced focal loss \cite{lin2017focal}:
\begin{gather}
D_f(\mathbf{x})=-\hat{\alpha}(1-\hat{p})^2\log(\hat{p}),\\
\hat{\alpha}=\begin{cases}
\alpha,&\text{if }y=1,\\
1-\alpha,&\text{otherwise},
\end{cases}\\
\hat{p}=\begin{cases}
p,&\text{if }y=1,\\
1-p,&\text{otherwise},
\end{cases}
\end{gather}
where $y\in\{0, 1\}$ is the ground-truth occupancy of a voxel $\mathbf{x}$ with $y=1$ indicating an occupied voxel, $p=f_{\mathbf{\Theta}}(\mathbf{x})$ is the predicted occupancy probability, and $\alpha\in(0,1)$ serves as a hyperparameter used to balance the two classes of voxels. We set $\alpha$ to be the proportion of the empty voxels.

\subsubsection{Attribute Distortion} The attribute distortion loss function per voxel is defined as
\begin{equation}
D_g(\widehat{\mathbf{x}})=\|\widehat{\mathbf{c}}-\mathbf{c_{\text{nearest}}}\|_2^2,
\end{equation}
where $\widehat{\mathbf{c}}=g_{\mathbf{\Phi}}(\widehat{\mathbf{x}})$ is the predicted attributes of the voxel $\widehat{\mathbf{x}}$, and $\mathbf{c}_{\text{nearest}}$ denotes the expected attribute, which is the ground-truth attribute of the voxel's nearest neighbor in the original point cloud.

\subsubsection{Sparsification Penalty}

Some existing works (e.g., \cite{hu2022nvfpcc, isik2022lvac}) utilize entropy models to estimate bit rates and minimize rate-distortion loss functions during training. However, we observe that employing entropy models can slow down and destabilize the training process. Inspired by $\ell_1$-regularization, we incorporate the $\ell_1$ norm of the network parameters into our loss functions. As $\ell_1$-regularization pushes network parameters toward zero, we can readily achieve sparse networks at low bit rates when employing DeepCABAC. Adjusting the regularization strengths $\lambda_f$ and $\lambda_g$ allows us to attain different bit rates.

\subsection{Sampling Strategy}

During the training of the network 
$f_{\mathbf{\Theta}}$, we randomly select voxels from $\mathcal{V}$ to form a training batch $\mathcal{B}_f$. It's worth noting that we control the ratio of occupied voxels within each batch. We represent this ratio with a specified hyperparameter $\beta\in(0,1)$. Recall that the focal loss uses a hyperparameter $\alpha$ to balance the samples, where $\alpha$ is the proportion of the empty voxels. Thus, we have $\alpha=1-\beta$.

\section{Experimental Results}\label{sec:III}
\begin{figure*}[t]
\centering
\includegraphics[width=2\columnwidth]{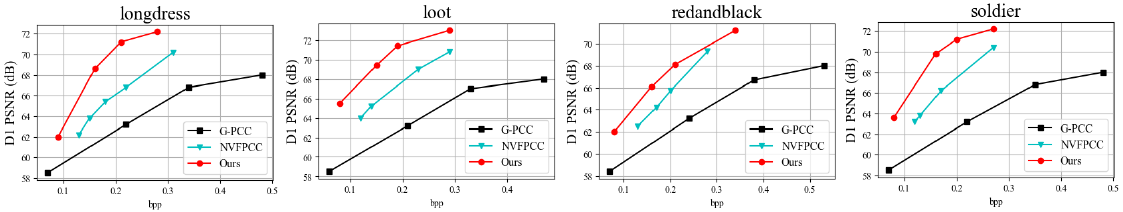}
\caption{Rate-distortion curves of different geometry compression methods, measured by D1 PSNR.}
\label{fig:geometry_result}
\end{figure*}

\begin{figure*}[t]
\centering
\includegraphics[width=2\columnwidth]{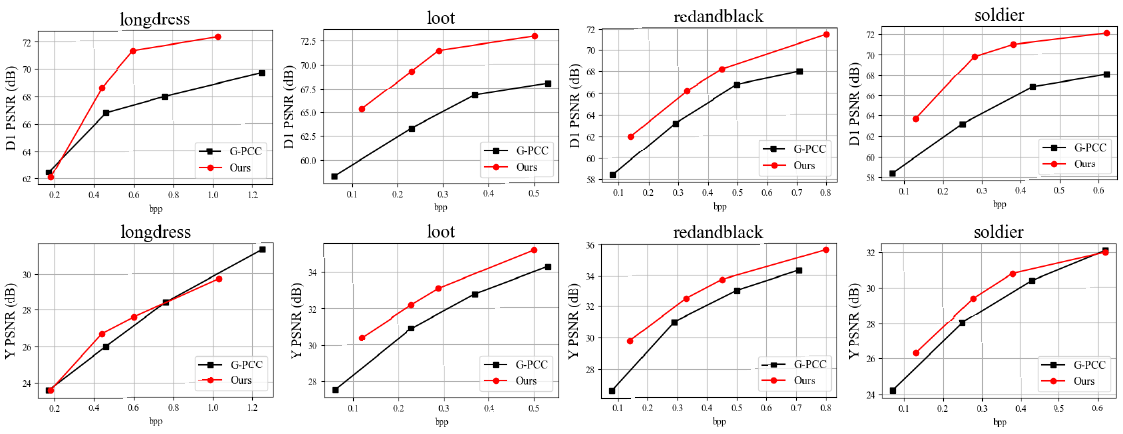}
\caption{Rate-distortion curves of the proposed method and G-PCC for joint geometry and attribute compression, measured by D1 PSNR and Y PSNR.}
\label{fig:joint_result}
\end{figure*}

\subsection{Experimental Settings}

We evaluate our proposed method on point clouds from 8i Voxelized Full Bodies (8iVFB) \cite{deon20178ivfb}, a dataset employed in MPEG PCC Common Testing Condition \cite{schwarz2018ctc}. Specifically, we used four point clouds in our evaluation, namely \textit{longdress}, \textit{loot}, \textit{redandblack}, and \textit{soldier}, all of which have been voxelized with a 10-bit resolution. We partition the space into $32\times 32\times 32$ cubes, i.e., $T=5$.

For each point cloud, we overfit the two networks $f_{\mathbf{\Theta}}$ and $g_{\mathbf{\Phi}}$whose detailed configurations are outlined in Table \ref{tab:network_configuration}. Both networks are trained using the Adam optimizer with a batch size of 4096 voxels. The first network, $f_{\mathbf{\Theta}}$, undergoes training for approximately 1200K steps, while the second network, $g_{\mathbf{\Phi}}$, is trained for about 200K steps. Initially, the learning rate is set to $10^{-3}$ and decays to $10^{-6}$ over the entire training process. During training of $f_{\mathbf{\Theta}}$, $\beta$ is set to 0.5, and the threshold $\tau$ is manually fine-tuned for optimal reconstruction quality. Additionally, we adjust $\lambda_f$ and $\lambda_g$ in the loss functions in order to achieve different bit rates.

\begin{table}
\centering
\caption{Parameter settings for networks $f_{\mathbf{\Theta}}$ and $g_{\mathbf{\Phi}}$.}
\label{tab:network_configuration}
\begin{tabular}{|c|c|c|}
\hline
Network & $f_{\mathbf{\Theta}}$ & $g_{\mathbf{\Phi}}$\\
\hline
Input channels & 3 & 3 \\
Output channels & 1 & 3 \\
Number of frequencies & 12 & 12 \\
Number of ResBlocks & 2 & 3 \\
ResBlock parameters & 512x128x512 & 512x128x512\\
Quantization step size & 1/1024 & 1/4096\\
\hline
\end{tabular}
\end{table}

Following the Common Test Condition \cite{schwarz2018ctc}, we quantify geometry distortion using point-to-point distance (D1) in peak signal-to-noise ratio (PSNR), while attribute distortion is assessed by Y PSNR. The compressed bit rate is defined by bits per point (bpp).

\subsection{Performance Evaluation}

We compare our method with the latest version of MPEG G-PCC reference software TMC13-v23.0-rc2 \cite{mpegpcctmc13}. We first consider the compression of only geometry components. We employ G-PCC (octree) as the benchmark for geometry compression. Additionally, we compare our method with NVFPCC \cite{hu2022nvfpcc}, an existing approach based on neural implicit representations. The rate-distortion performance comparison is illustrated in Fig. \ref{fig:geometry_result}. It can be observed that our method surpasses G-PCC (octree) and NVFPCC in terms of reconstruction quality for all four point clouds.

Then we compare the performance of the proposed approach on joint geometry and attribute compression to the latest G-PCC. Specifically, we employ G-PCC (octree) for geometry compression and G-PCC (RAHT) for attribute compression as benchmarks. We can see from Fig. \ref{fig:joint_result} that our method consistently outperforms G-PCC in terms of both geometry and attribute reconstruction quality for all four point clouds in most cases, validating the effectiveness of the proposed approach.

\begin{figure*}[t]
\centering
\includegraphics[width=2\columnwidth]{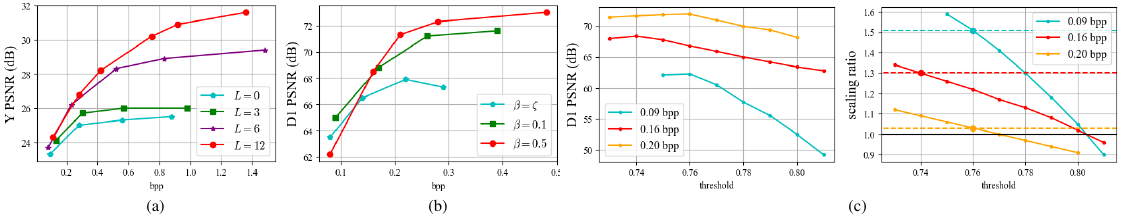}
\caption{Ablation studies on \textit{longdress}. (a) Performance of attribute compression with different $L$. (b) Performance of geometry compression when $\beta= 0.1, 0.5, \zeta$. (c) Reconstructed geometry distortion (Left) and scaling ratio (Right) with regard to threshold $\tau$ at different bit rates. The optimal ratios are indicated by dashed lines.}
\label{fig:ablation_studies}
\end{figure*}

\subsection{Ablation Studies}

In this section, we investigate the effectiveness of different modules in our proposed framework on the \textit{longdress} point cloud.

\subsubsection{Positional Encoding}

To demonstrate the benefits of the adpoted positional encoding given in Eq. (\ref{eq:positional_encoding}) and explore the effect of different $L$, we set $L$ to $0,3,6,12$ for the network $g_{\mathbf{\Phi}}$, resulting in four rate-distortion curves. Here, $L=0$ indicates that we do not apply positional encoding to the input voxel. The results are shown in Fig. \ref{fig:ablation_studies}a. We can see that positional encoding improves the network's ability to achieve high performance. It is evident that positional encoding enhances the network's capacity to achieve higher performance. Specifically, there is the minimal discrepancy in rate-distortion performance at lower bit rates. However, as the bit rate increases, networks with larger $L$ consistently outperform those with smaller $L$. Yet, beyond $L=12$, further increases in $L$ yield marginal performance improvements.

\subsubsection{Sampling Strategy}

As mentioned before, we introduce a hyperparameter $\beta$ controlling the proportion of occupied voxels in each training batch. We experiment with various values of $\beta$, i.e., $\beta=0.1, 0.5$, and $\zeta$. Here, $\beta=\zeta$ implies uniform voxel sampling across the volumetric space. The reconstruction performance with different $\beta$ values is depicted in Figure \ref{fig:ablation_studies}b. Notably, sampling more occupied voxels exhibits superior performance over uniform sampling, particularly at higher bit rates. Despite the focal loss also balancing occupied and empty voxels, experimental results reveal that sampling more occupied voxels further improves the training effectiveness.

\subsubsection{Occupancy Threshold}

Recall that in reconstructing voxel occupancies, we introduce a threshold $\tau$ to determine whether a voxel is occupied or not. To assess the impact of this threshold, we first train a network $f_{\mathbf{\Theta}}$ for each bit rate and then experiment with different values of $\tau$. The results are presented in Figure \ref{fig:ablation_studies}c (Left). Notably, the reconstruction performance of our approach varies significantly with the threshold value, and the optimal threshold minimizing distortion varies across different bit rates. Since the threshold influences the number of reconstructed points (i.e., $|\widehat{\mathcal{X}}|$), we compare it with the ground-truth value $|\mathcal{X}|$ by computing the scaling ratio $|\widehat{\mathcal{X}}|/|\mathcal{X}|$. Figure \ref{fig:ablation_studies}c (Right) illustrates that the scaling ratio decreases with increasing threshold values. Dashed lines indicate the optimal ratios minimizing D1 PSNR distortion. Interestingly, these optimal ratios decrease as bit rates increase. Ideally, the scaling ratio should approach 1 for high reconstruction quality. However, in practice, due to the definition of D1 PSNR metrics, optimal ratios are typically greater than 1, especially at low bit rates.

\subsection{Discussion on Universality}

\begin{figure*}[t]
\centering
\includegraphics[height=4cm]{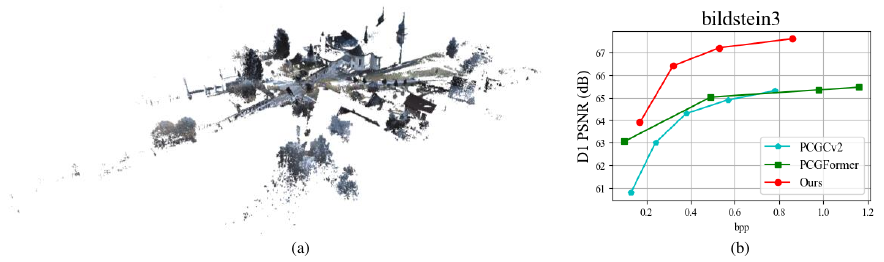}
\caption{Experiments on \textit{bildstein3}. (a) The original point cloud (voxelized with 11-bit resolution). (b) Rate-distortion curves by different geometry compression methods, measured by D1 PSNR.}
\label{fig:bildstein3}
\end{figure*}


Existing learning-based point cloud compression methods \cite{wang2021pcgcv2, liu2022pcgformer, wang2022sparsepcgc, wang2022sparsepcac} typically train a model on a large dataset of point clouds, achieving superior performance on clouds similar to those in the training set. However, these methods may suffer performance degradation on out-of-distribution point clouds, indicating a lack of generalization. In contrast, our proposed approach customizes two neural networks for each individual point cloud, thus not confined to a specific training dataset and applicable to any type of point cloud.

To demonstrate the universality of our method, we select the point cloud \textit{bildstein3} from semantic3d.net \cite{hackel2017semantic3d}, a dataset of urban and rural scenes. We manually voxelize \textit{bildstein3} with an 11-bit resolution, shown in Fig. \ref{fig:bildstein3}a. We compare the rate-distortion performance of the proposed approach with that of two learning-based benchmarks, i.e., PCGCv2 \cite{wang2021pcgcv2} and PCGFormer \cite{liu2022pcgformer}. As shown in Fig. \ref{fig:bildstein3}b, our method significantly outperforms PCGCv2 and PCGFormer, especially at high bit rates. This may be because \textit{bildstein3} contains many scattered points in the volumetric space, making it notably different from the point clouds used for training the benchmarks.

\section{Conclusion}\label{sec:IV}
This paper introduced a novel approach for point cloud compression based on neural implicit representation. Specifically, we utilize two neural networks to implicitly encode geometry and attribute information for each point cloud, followed by parameter quantization and encoding. Our experimental results demonstrated that our method surpasses the latest G-PCC in terms of rate-distortion performance. Furthermore, when applying geometry compression methods to entirely different point clouds, our approach exhibited significant performance gains over existing learning-based methods, showcasing its universality. This work opens new avenues and opportunities for advancing point cloud compression techniques, with considerable potential for future research to enhance rate-distortion performance. Additionally, our approach shows promise for extension to dynamic point cloud compression, as neural implicit representations can effectively leverage shared information across frames.

\small
\bibliographystyle{IEEEtran}
\bibliography{ref.bib}

\begin{thebibliography}{10}
\providecommand{\url}[1]{#1}
\csname url@samestyle\endcsname
\providecommand{\newblock}{\relax}
\providecommand{\bibinfo}[2]{#2}
\providecommand{\BIBentrySTDinterwordspacing}{\spaceskip=0pt\relax}
\providecommand{\BIBentryALTinterwordstretchfactor}{4}
\providecommand{\BIBentryALTinterwordspacing}{\spaceskip=\fontdimen2\font plus
\BIBentryALTinterwordstretchfactor\fontdimen3\font minus \fontdimen4\font\relax}
\providecommand{\BIBforeignlanguage}[2]{{%
\expandafter\ifx\csname l@#1\endcsname\relax
\typeout{** WARNING: IEEEtran.bst: No hyphenation pattern has been}%
\typeout{** loaded for the language `#1'. Using the pattern for}%
\typeout{** the default language instead.}%
\else
\language=\csname l@#1\endcsname
\fi
#2}}
\providecommand{\BIBdecl}{\relax}
\BIBdecl

\bibitem{graziosi2020pccstd}
D.~Graziosi, O.~Nakagami, S.~Kuma, A.~Zaghetto, T.~Suzuki, and A.~Tabatabai, ``An overview of ongoing point cloud compression standardization activities: Video-based (v-pcc) and geometry-based (g-pcc),'' \emph{APSIPA Transactions on Signal and Information Processing}, vol.~9, p. e13, 2020.

\bibitem{wang2021learnedpcgc}
J.~Wang, H.~Zhu, H.~Liu, and Z.~Ma, ``Lossy point cloud geometry compression via end-to-end learning,'' \emph{IEEE Transactions on Circuits and Systems for Video Technology}, vol.~31, no.~12, pp. 4909--4923, 2021.

\bibitem{PCmag}
Y.~Shao, C.~Bian, L.~Yang, Q.~Yang, Z.~Zhang, and D.~Gunduz, ``Point cloud in the air,'' \emph{arXiv:2401.00658}, 2024.

\bibitem{wang2021pcgcv2}
J.~Wang, D.~Ding, Z.~Li, and Z.~Ma, ``Multiscale point cloud geometry compression,'' in \emph{2021 Data Compression Conference (DCC)}.\hskip 1em plus 0.5em minus 0.4em\relax IEEE, 2021, pp. 73--82.

\bibitem{liu2022pcgformer}
G.~Liu, J.~Wang, D.~Ding, and Z.~Ma, ``Pcgformer: Lossy point cloud geometry compression via local self-attention,'' in \emph{2022 IEEE International Conference on Visual Communications and Image Processing (VCIP)}.\hskip 1em plus 0.5em minus 0.4em\relax IEEE, 2022, pp. 1--5.

\bibitem{wang2022sparsepcgc}
J.~Wang, D.~Ding, Z.~Li, X.~Feng, C.~Cao, and Z.~Ma, ``Sparse tensor-based multiscale representation for point cloud geometry compression,'' \emph{IEEE Transactions on Pattern Analysis and Machine Intelligence}, 2022.

\bibitem{wang2022sparsepcac}
J.~Wang and Z.~Ma, ``Sparse tensor-based point cloud attribute compression,'' in \emph{2022 IEEE 5th International Conference on Multimedia Information Processing and Retrieval (MIPR)}.\hskip 1em plus 0.5em minus 0.4em\relax IEEE, 2022, pp. 59--64.

\bibitem{SEPT}
C.~Bian, Y.~Shao, and D.~Gunduz, ``Wireless point cloud transmission,'' \emph{arXiv preprint:2306.08730}, 2023.

\bibitem{alexiou2020unified}
E.~Alexiou, K.~Tung, and T.~Ebrahimi, ``Towards neural network approaches for point cloud compression,'' in \emph{Applications of digital image processing XLIII}, vol. 11510.\hskip 1em plus 0.5em minus 0.4em\relax SPIE, 2020, pp. 18--37.

\bibitem{mildenhall2021nerf}
B.~Mildenhall, P.~P. Srinivasan, M.~Tancik, J.~T. Barron, R.~Ramamoorthi, and R.~Ng, ``Nerf: Representing scenes as neural radiance fields for view synthesis,'' \emph{Communications of the ACM}, vol.~65, no.~1, pp. 99--106, 2021.

\bibitem{hu2022nvfpcc}
Y.~Hu and Y.~Wang, ``Learning neural volumetric field for point cloud geometry compression,'' in \emph{2022 Picture Coding Symposium (PCS)}.\hskip 1em plus 0.5em minus 0.4em\relax IEEE, 2022, pp. 127--131.

\bibitem{isik2022lvac}
B.~Isik, P.~A. Chou, S.~J. Hwang, N.~Johnston, and G.~Toderici, ``Lvac: Learned volumetric attribute compression for point clouds using coordinate based networks,'' \emph{Frontiers in Signal Processing}, vol.~2, p. 1008812, 2022.

\bibitem{pistilli2022nic}
F.~Pistilli, D.~Valsesia, G.~Fracastoro, and E.~Magli, ``Signal compression via neural implicit representations,'' in \emph{ICASSP 2022-2022 IEEE International Conference on Acoustics, Speech and Signal Processing (ICASSP)}.\hskip 1em plus 0.5em minus 0.4em\relax IEEE, 2022, pp. 3733--3737.

\bibitem{wiedemann2020deepcabac}
S.~Wiedemann, H.~Kirchhoffer, S.~Matlage, P.~Haase, A.~Marban, T.~Marin{\v{c}}, D.~Neumann, T.~Nguyen, H.~Schwarz, T.~Wiegand \emph{et~al.}, ``Deepcabac: A universal compression algorithm for deep neural networks,'' \emph{IEEE Journal of Selected Topics in Signal Processing}, vol.~14, no.~4, pp. 700--714, 2020.

\bibitem{he2016residual}
K.~He, X.~Zhang, S.~Ren, and J.~Sun, ``Deep residual learning for image recognition,'' in \emph{Proceedings of the IEEE conference on computer vision and pattern recognition}, 2016, pp. 770--778.

\bibitem{lin2017focal}
T.-Y. Lin, P.~Goyal, R.~Girshick, K.~He, and P.~Doll{\'a}r, ``Focal loss for dense object detection,'' in \emph{Proceedings of the IEEE international conference on computer vision}, 2017, pp. 2980--2988.

\bibitem{deon20178ivfb}
E.~d’Eon, B.~Harrison, T.~Myers, and P.~A. Chou, ``8i voxelized full bodies-a voxelized point cloud dataset,'' \emph{ISO/IEC JTC1/SC29 Joint WG11/WG1 (MPEG/JPEG) input document WG11M40059/WG1M74006}, vol.~7, no.~8, p.~11, 2017.

\bibitem{schwarz2018ctc}
S.~Schwarz, G.~Martin-Cocher, D.~Flynn, and M.~Budagavi, ``Common test conditions for point cloud compression,'' \emph{Document ISO/IEC JTC1/SC29/WG11 w17766, Ljubljana, Slovenia}, 2018.

\bibitem{mpegpcctmc13}
\BIBentryALTinterwordspacing
MPEGGroup. Mpeg-pcc-tmc13. (2024, Feb 7). [Online]. Available: \url{https://github.com/MPEGGroup/mpeg-pcc-tmc13}
\BIBentrySTDinterwordspacing

\bibitem{hackel2017semantic3d}
T.~Hackel, N.~Savinov, L.~Ladicky, J.~D. Wegner, K.~Schindler, and M.~Pollefeys, ``Semantic3d. net: A new large-scale point cloud classification benchmark,'' \emph{arXiv:1704.03847}, 2017.

\end{thebibliography}

\end{document}